\documentclass[conference,a4paper]{IEEEtran}
\IEEEoverridecommandlockouts

\usepackage{graphicx}
\usepackage{subcaption}
\usepackage{hyperref}
\usepackage{booktabs}
\usepackage{xcolor}
\usepackage[T1]{fontenc}
\usepackage[absolute,overlay]{textpos}

\begin{document}

\title{Toward AI-Assisted Poultry Coccidiosis Diagnosis: Evaluating Gemini and BiomedParse on Eimeria Microscopy Images}

\author{\IEEEauthorblockN{Ali Alsalama\IEEEauthorrefmark{1},
Ahmed Kubba\IEEEauthorrefmark{1},
Manar Abu Talib\IEEEauthorrefmark{1}
}
\IEEEauthorblockA{\IEEEauthorrefmark{1}Department of Computer Science, \\
College of Computing and Informatics,\\
University of Sharjah, \\
Sharjah, United Arab Emirates\\
Email: \{U23102893, u23103280, mtalib\}@sharjah.ac.ae}
}
\maketitle

% ===== First-page conference header and IEEE copyright footer (page 1 only) =====
\begin{textblock*}{\textwidth}(\dimexpr(\paperwidth-\textwidth)/2\relax,0.4in)
\centering\footnotesize 2026 International Conference on Sustainability, Innovation \& Technology (ICSIT)
\end{textblock*}
\begin{textblock*}{\textwidth}(\dimexpr(\paperwidth-\textwidth)/2\relax,\dimexpr\paperheight-0.55in\relax)
{\footnotesize 979-8-3315-4827-8/26/\$31.00~\copyright2026 IEEE}
\end{textblock*}
% ============================================================================

\begin{abstract}
Coccidiosis caused by \textit{Eimeria} parasites is a major economic burden in poultry production, and effective control depends on accurate species-level diagnosis. This study evaluates whether a general-purpose multimodal large language model can support such diagnosis. Google Gemini was assessed on 4,225 microscopy images covering the seven fowl-infecting \textit{Eimeria} species under two prompting conditions, one without candidate labels and one with a predefined class list, and was further tested for pathology-report generation, while BiomedParse was examined for parasite segmentation. Without candidate labels, the model produced broad and taxonomically inconsistent outputs. With candidate labels, overall accuracy reached only 14.9\%, with a strong bias toward \textit{E. tenella} at 74\% and no correct classifications for \textit{E. acervulina}, \textit{E. mitis} and \textit{E. praecox}. Generated treatment reports were coherent but unverified, and segmentation was only partial. Current multimodal models are therefore not yet reliable for standalone \textit{Eimeria} diagnosis without domain-specific fine-tuning and expert validation.
\end{abstract}


\section{Introduction}
\textit{Eimeria} parasites are a major concern in poultry farming, responsible for coccidiosis, a disease characterized by intestinal damage, reduced feed efficiency, and, in severe cases, high mortality \cite{LopezOsorio2020}. Accurate species-level identification is critical for targeted treatment with vaccines and anticoccidial drugs, yet conventional diagnosis by microscopy and molecular methods is labor-intensive and requires specialized expertise \cite{Marugan2020}. Recent advances in artificial intelligence (AI), particularly large language models (LLMs) able to process both textual and visual inputs, offer new opportunities to automate and support this process \cite{Kasneci2023,Chang2024}.

The genus \textit{Eimeria} comprises seven species that infect chickens, each differing in pathogenicity and in the gastrointestinal region it targets. \textit{E. tenella} and \textit{E. necatrix} are highly pathogenic, causing severe hemorrhagic lesions and high mortality, whereas \textit{E. acervulina} and \textit{E. maxima} cause moderate disease with reduced growth and feed efficiency, and \textit{E. brunetti}, \textit{E. mitis}, and \textit{E. praecox} range from moderate to largely subclinical infections \cite{Burrell2020,Schnitzler1999,Xu2022}. Their distinct oocyst morphology makes accurate classification important for effective coccidiosis management.

While LLMs and their multimodal variants have advanced rapidly in natural-language and medical-imaging tasks \cite{Teubner2023}, their application to specialized veterinary parasitology remains largely unexplored, and their ability to identify \textit{Eimeria} species and generate pathology reports from microscopic images has not been systematically evaluated. This gap is significant given the rise of drug-resistant strains and the need for scalable diagnostics in resource-constrained settings.

This study evaluates the feasibility of using multimodal LLMs, focusing on Google Gemini, to support the diagnosis and treatment interpretation of \textit{Eimeria} infections from microscopic images, using an \textit{Eimeria} fowl dataset of 4,225 images across the seven species. Gemini is assessed under different prompting settings to examine its ability to identify parasite classes, discriminate among species when candidate labels are provided, and generate pathology-style reports with suggested treatments. Additionally, BiomedParse is explored as a biomedical foundation model for parasite segmentation in future work. The code used in this study is available at \href{https://github.com/alialsalamaa/eimeria-classification-with-gemini}{the project repository}. The specific objectives are: (i)~to evaluate Gemini's classification of \textit{Eimeria} microscopic images with and without candidate species labels, (ii)~to analyze its class-wise performance, prediction bias, and ability to generate treatment-oriented pathology reports, and (iii)~to assess BiomedParse for \textit{Eimeria} segmentation and identify the requirements for deploying such tools in practice.

\section{Literature Review}
The accurate identification of \textit{Eimeria} species is considered essential for appropriate treatment, but traditional morphology- and molecular-based diagnosis tend to be time-consuming, expertise-dependent, and difficult to scale, motivating the development of automated AI-based image analysis techniques \cite{xiao2025}.

Machine learning and deep learning are increasingly applied across veterinary diagnostic imaging, disease classification, and clinical decision support \cite{perezgarcia2026,huang2026}. In parallel, multimodal LLMs (MLLMs) combine visual interpretation with language reasoning and have shown rapid progress in radiology report generation, medical visual question answering, and diagnostic assistance \cite{yi2025,nam2025,xiao2026}, including dedicated models that generate structured reports from images and whole-slide pathology \cite{li2025,hu2025}. Recent studies have also started to evaluate LLMs in veterinary settings, including undergraduate examinations \cite{alonsosousa2025}, feline ocular and fundus diagnosis \cite{okur2025,eren2025}, clinical-record summarization \cite{poore2025}, and knowledge-graph-augmented bovine diagnosis \cite{qu2026}. Together, these studies show that general-purpose LLMs can support veterinary reasoning but require domain grounding and rigorous, task-specific evaluation.

A substantial body of work has specifically targeted automated classification of the seven fowl-infecting \textit{Eimeria} species, which provides a direct basis for comparison. Casta\~{n}\'{o}n et al. used hand-crafted oocyst shape descriptors and reported 85.75\% accuracy on the University of S\~{a}o Paulo (COCCIMORPH) \textit{Eimeria} image database \cite{Castanon2007}. CNN-based methods improved this to 90.42\% \cite{Monge2019}, and He et al.'s ResTFG, a lightweight CNN--Transformer hybrid, reached 96.9\% accuracy with only 1.95M parameters at 256 frames per second \cite{He2023}; further deep-learning approaches include neutrosophic-set preprocessing \cite{Sayed2024} and automated coccidia-detection pipelines \cite{Kellogg2024}. These specialized models achieve reliable discrimination but each requires curated training data and per-task optimization. In contrast, the present study evaluates a general-purpose MLLM (Google Gemini) off the shelf, without task-specific fine-tuning, and additionally probes report-generation and segmentation behavior that conventional classifiers do not provide. The application of MLLMs to \textit{Eimeria} microscopy has not previously been examined.

\section{Methodology}
\subsection{Eimeria Fowl Dataset}
This study uses an \textit{Eimeria} parasite image database for assessing the performance of LLMs, which consists of seven fowl-infecting \textit{Eimeria} species and contains a total of 4,225 images distributed among the different classes \cite{Acmali2024}. The list of classes, representing different \textit{Eimeria} species, alongside their respective number of images in the dataset can be observed in Table~\ref{tab:eimeria}.

The images originate from the publicly available COCCIMORPH \textit{Eimeria} image database at the University of S\~{a}o Paulo \cite{Castanon2007}, where each sample is a brightfield light-microscopy micrograph of a single sporulated oocyst. The images are RGB JPEG files cropped to one oocyst, with small, non-uniform dimensions (roughly 200--490 px per side, most 250--350 px), and were not captured under a standardized protocol: illumination, focus, and background vary, no common scale bar or acquisition metadata is provided, and the class distribution is imbalanced (Table~\ref{tab:eimeria}). These properties, namely the low and variable resolution, JPEG compression, and heterogeneous capture conditions, limit the fine morphological detail available to any image-based model and affect reproducibility. All experiments use the images as provided, without resizing, color normalization, or enhancement.

\begin{table}[htbp]
\caption{Image Distribution Across \textit{Eimeria} Parasite Classes}
\label{tab:eimeria}
\centering
\renewcommand{\arraystretch}{1.15}
\begin{tabular}{lc}
\hline
\textbf{Class} & \textbf{Image Count} \\
\hline
\textit{E. acervulina} & 726 \\
\textit{E. brunetti}   & 435 \\
\textit{E. maxima}     & 339 \\
\textit{E. mitis}      & 795 \\
\textit{E. necatrix}   & 468 \\
\textit{E. praecox}    & 857 \\
\textit{E. tenella}    & 605 \\
\hline
\end{tabular}
\end{table}
\subsection{Gemini Model}
Gemini is a multimodal model family from Google DeepMind that integrates vision-language capabilities \cite{Imran2024}. Gemini 1.5, the most advanced version available at the time of the experiments, was evaluated over the entire dataset using the Gemini API within a Google Colab environment. A processing function iterates over the dataset images and queries the model with the three prompts below. The first two prompts request one-word answers to enable reliable automated parsing of large batches, and a retry mechanism with exponential backoff (up to six attempts per image) handles temporary API failures.

\begin{itemize}
    \item \textbf{First Prompt:} ``This image belongs to a parasite, can you identify which class of parasite it belongs to? Answer in one word only.''
    \item \textbf{Second Prompt:} ``This may be a parasite in a chicken, can you classify this Eimeria type? It is one of these possible classes: ACERVULINA, BRUNETTI, MAXIMA, MITIS, NECATRIX, PRAECOX, or TENELLA. Answer in one word only.''
    \item \textbf{Third Prompt:} ``This is an image of a/an [parasite class] parasite in a chicken. Provide a pathology report which includes which antibiotic is needed and in what quantity to cure it.''
\end{itemize}

These prompts test complementary capabilities under increasing guidance. The first is open-ended and withholds candidate labels, testing unconstrained recognition and revealing the model's prior rather than forcing a choice. The second constrains the output to the seven candidate species, isolating fine-grained discrimination within the correct hypothesis space, yielding comparable per-class accuracy, precision, recall, and F1, and mirroring a realistic decision-support setting in which the diagnosis is already narrowed to poultry \textit{Eimeria}. The third supplies the ground-truth class and requests a pathology report, separating report generation from classification.

\subsection{BiomedParse Model}
BiomedParse is a cutting-edge biomedical foundation model designed to perform segmentation, detection, and recognition across nine imaging modalities. Using joint learning, it enhances task-specific accuracy and enables innovative applications, such as segmenting all relevant objects in an image using textual descriptions. Trained on over 6 million image, segmentation mask, and textual description triples, BiomedParse utilizes natural language labels from existing datasets. It surpasses previous methods in image segmentation, particularly excelling with irregularly shaped objects, and can simultaneously segment and label all objects in each given image \cite{Zhao2024}.

BiomedParse was tested on \textit{Eimeria} images from the dataset to assess its capability to segment \textit{Eimeria} parasite images. Even though this model was trained for human medical segmentation, testing its capabilities on \textit{Eimeria} images can still provide insights into possible fine-tuning and utilization of the model for the segmentation of chicken parasites.

\section{Results and Discussion}
\subsection{Parasite Classification}
\subsubsection{No initial information given in the input prompt}
A sample of the Gemini model's predictions for each class, across different iterations and images, when given no initial information to assist in its prediction can be observed in Table~\ref{tab:gemini_predictions}.

\begin{table}[htbp]
\caption{Gemini Predictions for the Input Prompt}
\label{tab:gemini_predictions}
\centering
\footnotesize
\begin{tabular}{p{1.7cm} p{5.0cm}}
\hline
\textbf{Class} & \textbf{Gemini Predictions} \\ \hline
\textit{E. acervulina} & Protozoa, Fungi \\
\textit{E. brunetti}   & Helminth, Protozoa \\
\textit{E. maxima}     & Helminth, Coccidia \\
\textit{E. mitis}      & Apicomplexa, Coccidia, Trematoda, Ciliophora \\
\textit{E. necatrix}   & Trematoda, Coccidia, Ciliata, Sporozoa, Apicomplexa \\
\textit{E. praecox}    & Helminth \\
\textit{E. tenella}    & Helminth, Nematode \\ \hline
\end{tabular}
\end{table}

Without candidate labels the predictions are broad and taxonomically inconsistent, reflecting the difficulty of identifying \textit{Eimeria} at the species level and the strong overlap in visual features across parasite groups.

\subsubsection{List of classes given in the input prompt}
Based on the responses of the Gemini model, the accuracy of the total predictions for each class can be observed in Table~\ref{tab:gemini_accuracy}.

\begin{table}[htbp]
\caption{Accuracy of Gemini predictions for the prompt: ``This may be a parasite in a chicken, can you classify this Eimeria type? It is one of these possible classes: ACERVULINA, BRUNETTI, MAXIMA, MITIS, NECATRIX, PRAECOX, or TENELLA. Answer in one word only.''}
\label{tab:gemini_accuracy}
\setlength{\tabcolsep}{9pt}
\renewcommand{\arraystretch}{1.1}
\begin{tabular}{lcccc}
\hline
\textbf{Class} & \textbf{Acc.} & \textbf{Prec.} & \textbf{Rec.} & \textbf{F1} \\
\hline
\textit{E. acervulina} & 0\%   & 0.000 & 0.000 & 0.000 \\
\textit{E. brunetti}   & 3\%   & 0.091 & 0.029 & 0.044 \\
\textit{E. maxima}     & 1.4\% & 0.030 & 0.014 & 0.019 \\
\textit{E. mitis}      & 0\%   & 0.000 & 0.000 & 0.000 \\
\textit{E. necatrix}   & 14\%  & 0.094 & 0.143 & 0.114 \\
\textit{E. praecox}    & 0\%   & 0.000 & 0.000 & 0.000 \\
\textit{E. tenella}    & 74\%  & 0.174 & 0.756 & 0.283 \\
\hline
\textbf{Macro avg} & -- & \textbf{0.056} & \textbf{0.135} & \textbf{0.066} \\
\textbf{Overall acc.} & \textbf{14.9\%} & \multicolumn{3}{c}{\textbf{0.149}} \\
\hline
\end{tabular}
\end{table}

\begin{figure}[htbp]
    \centering
    \includegraphics[width=0.48\textwidth]{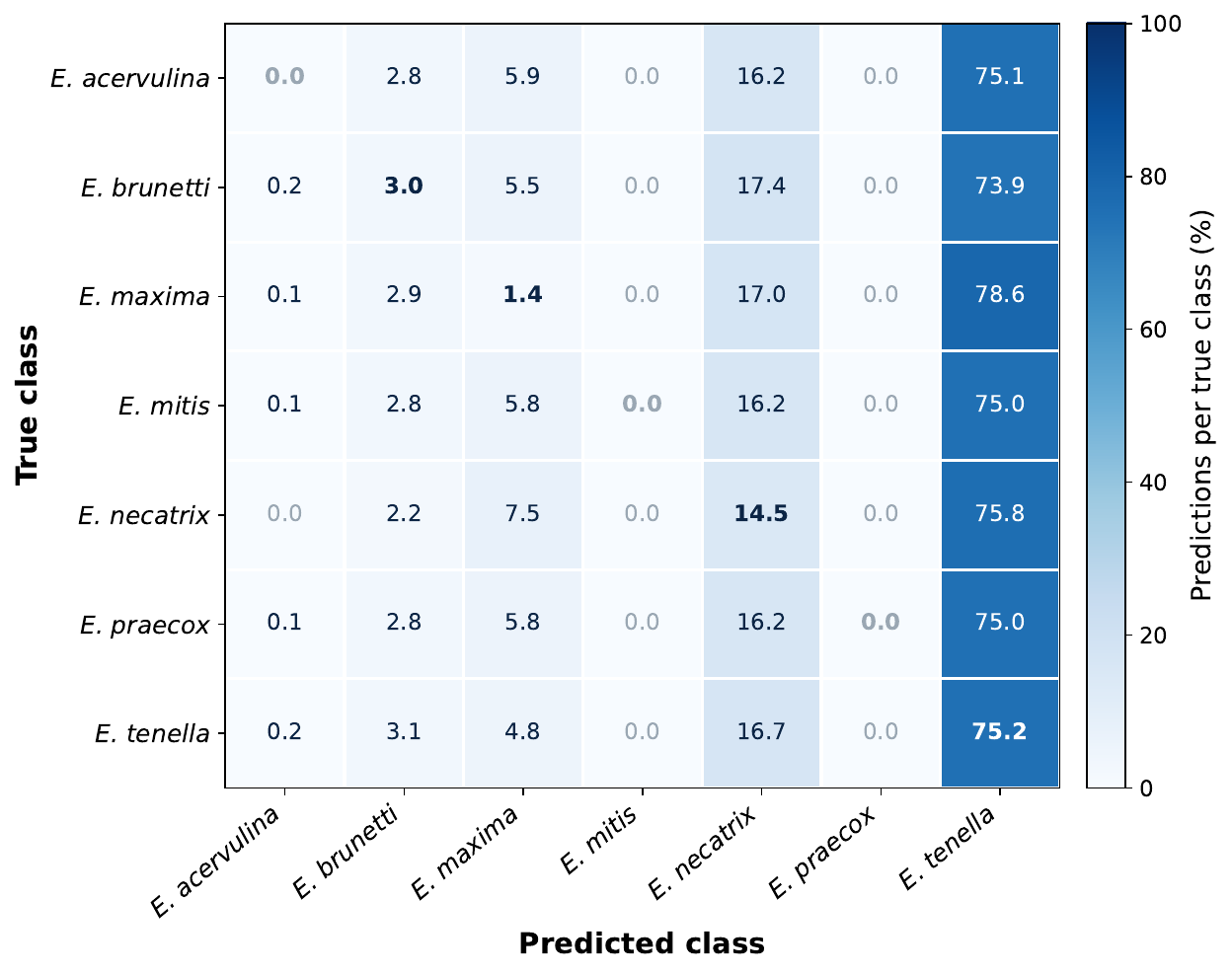}
    \caption{Row-normalized confusion matrix (percentage of each true class) for the constrained candidate-label prompt. The bright column under \textit{E. tenella} shows that most images of every species are predicted as \textit{E. tenella}, confirming the strong prediction bias, while the all-zero \textit{E. mitis} and \textit{E. praecox} columns indicate that the model never assigned those labels.}
    \label{fig:confusion}
\end{figure}

To visualize the class bias summarized in Table~\ref{tab:gemini_accuracy}, Fig.~\ref{fig:confusion} presents the row-normalized confusion matrix for the constrained candidate-label prompt. Each row reports the percentage of images assigned to each predicted species for a given true species, so the diagonal corresponds to the per-class recall. The matrix makes the bias clearly observable, as the model predicts \textit{E. tenella} for roughly three quarters of the images in every species, whereas the \textit{E. mitis} and \textit{E. praecox} labels are essentially never produced.

These results show highly inconsistent performance across species. The 74\% accuracy for \textit{E. tenella} does not reflect genuine recognition. As Fig.~\ref{fig:confusion} shows, the model labels roughly three quarters of the images in \emph{every} species as \textit{E. tenella}, so this score largely reflects a default prediction. \textit{E. necatrix} reaches only 14\%, \textit{E. brunetti} and \textit{E. maxima} are at 3\% and 1.4\%, and \textit{E. acervulina}, \textit{E. mitis}, and \textit{E. praecox} are never correctly classified. This bias likely reflects uneven representation of these species in the model's pretraining, together with the subtle morphological differences between them. Task-specific fine-tuning, balanced and augmented data, and domain-informed prompting would be required to improve performance.

\subsection{Pathology Report Generation}
\begin{figure*}[!htbp]
\centering
\begin{minipage}[t]{0.31\textwidth}
    \centering
    \includegraphics[width=3.2cm]{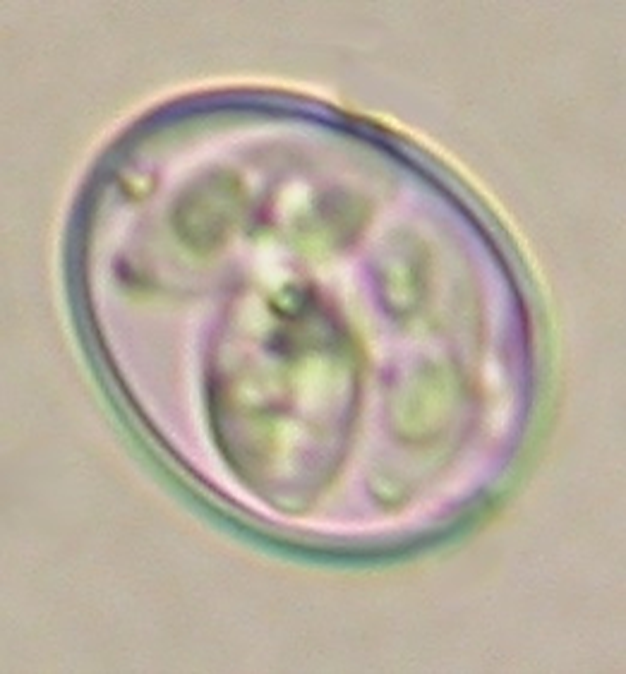}\\[0.25cm]
    \footnotesize\raggedright
    \textbf{Input Prompt:} This is an image of an \textit{E.~acervulina} parasite in a chicken. Provide a pathology report which includes which antibiotic is needed and in what quantity to cure it.\\[0.15cm]
    \textbf{Antibiotics:}\\
    \textbf{Amprolium:} Prevention 125~ppm in feed continuously; treatment 250~ppm in drinking water for 5--7 days.\\
    \textbf{Sulfaquinoxaline:} Prevention 0.05\% in feed for 2--3 weeks; treatment 0.1\% in drinking water for 2 days, then 0.05\% for 4 days.
\end{minipage}\hfill
\begin{minipage}[t]{0.31\textwidth}
    \centering
    \includegraphics[width=3.2cm]{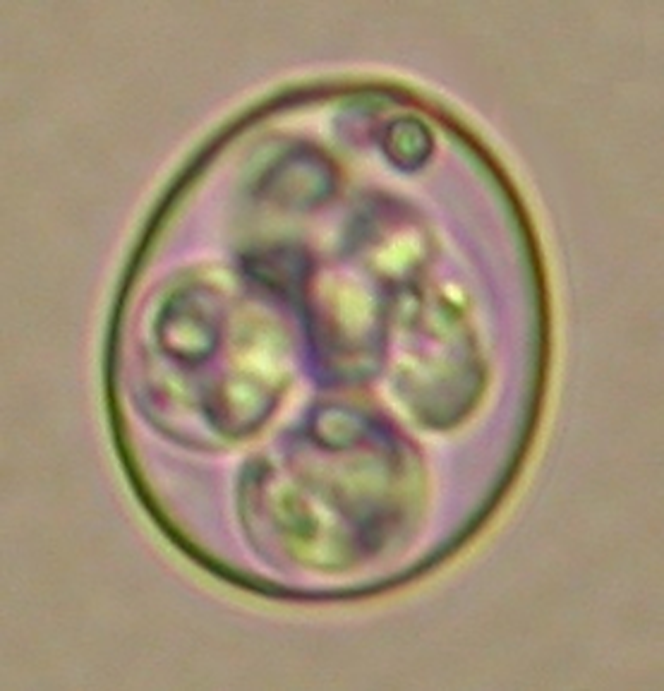}\\[0.25cm]
    \footnotesize\raggedright
    \textbf{Input Prompt:} This is an image of an \textit{E.~brunetti} parasite in a chicken. Provide a pathology report which includes which antibiotic is needed and in what quantity to cure it.\\[0.15cm]
    \textbf{Antibiotics:}\\
    \textbf{Sulfaquinoxaline:} Prevention 0.05\% in feed for 2--3 weeks; treatment 0.1\% in drinking water for 2 days, then 0.05\% for 4 days.\\
    \textbf{Amprolium:} Prevention 125~ppm in feed continuously; treatment 250~ppm in drinking water for 5--7 days.\\
    \textbf{Toltrazuril:} Consult a veterinarian for appropriate dosage.
\end{minipage}\hfill
\begin{minipage}[t]{0.31\textwidth}
    \centering
    \includegraphics[width=3.2cm]{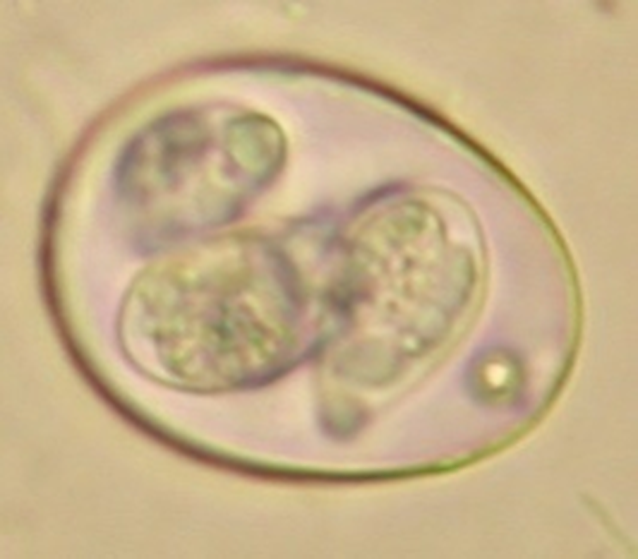}\\[0.25cm]
    \footnotesize\raggedright
    \textbf{Input Prompt:} This is an image of an \textit{E.~maxima} parasite in a chicken. Provide a pathology report which includes which antibiotic is needed and in what quantity to cure it.\\[0.15cm]
    \textbf{Antibiotics:}\\
    \textbf{Sulfaquinoxaline:} Prevention 0.05\% in feed for 2--3 weeks; treatment 0.1\% in drinking water for 2 days, then 0.05\% for 4 days.\\
    \textbf{Amprolium:} Prevention 125~ppm in feed continuously; treatment 250~ppm in drinking water for 5--7 days.\\
    \textbf{Toltrazuril:} Consult a veterinarian for appropriate dosage.
\end{minipage}
\caption{Sample pathology reports generated by the Gemini model for selected \textit{Eimeria} parasite classes. Each panel shows the input oocyst image, the input prompt, and the treatment recommendation returned by the model.}
\label{fig:parasite_reports_first3}
\end{figure*}

The pathology reports generated by the Gemini model for given \textit{Eimeria} parasite classes are shown in Fig.~\ref{fig:parasite_reports_first3}, which presents the input image alongside the prompt and the treatment plan and antibiotics suggested by the model.

The reports are coherent and clinically plausible, and the suggested anticoccidials (amprolium, sulfaquinoxaline, toltrazuril) and their dosages broadly match common coccidiosis protocols. They were not, however, verified by a veterinary expert, which is an essential step before any practical use. LLMs can produce confident but incorrect details, and appropriate drug choice and dosing depend on factors a general-purpose model may not capture, such as region-specific approvals and withdrawal periods, local anticoccidial resistance, and flock age and weight. A rigorous assessment would require board-certified parasitologists to rate each report for factual accuracy, dosage correctness, and safety against established formularies; such expert validation is a prerequisite for clinical decision support and a key direction for future work.

\subsection{BiomedParse Segmentation}
\begin{figure}[htbp]
\centering
\begin{subfigure}{0.92\columnwidth}
    \includegraphics[width=\textwidth]{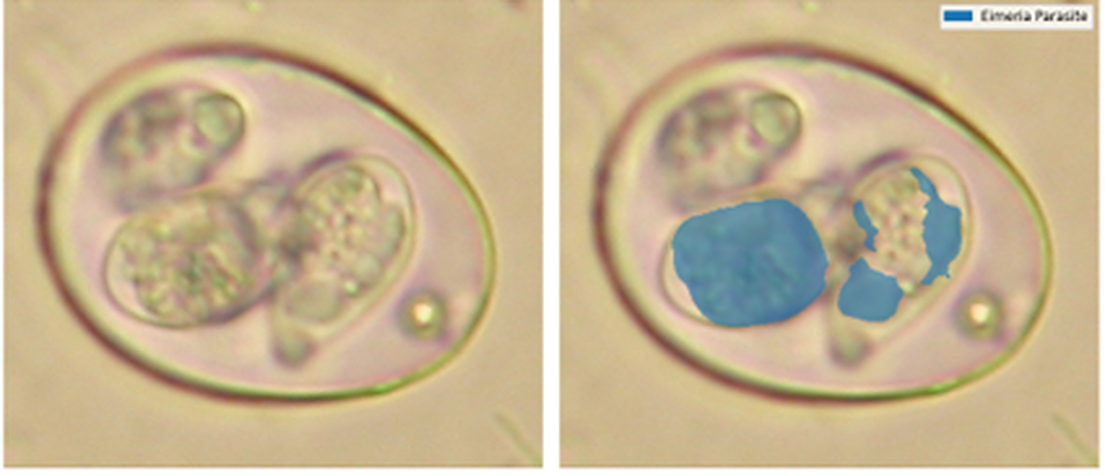}
    \caption{\textit{E. maxima}}
\end{subfigure}

\vspace{0.15cm}
\begin{subfigure}{0.92\columnwidth}
    \includegraphics[width=\textwidth]{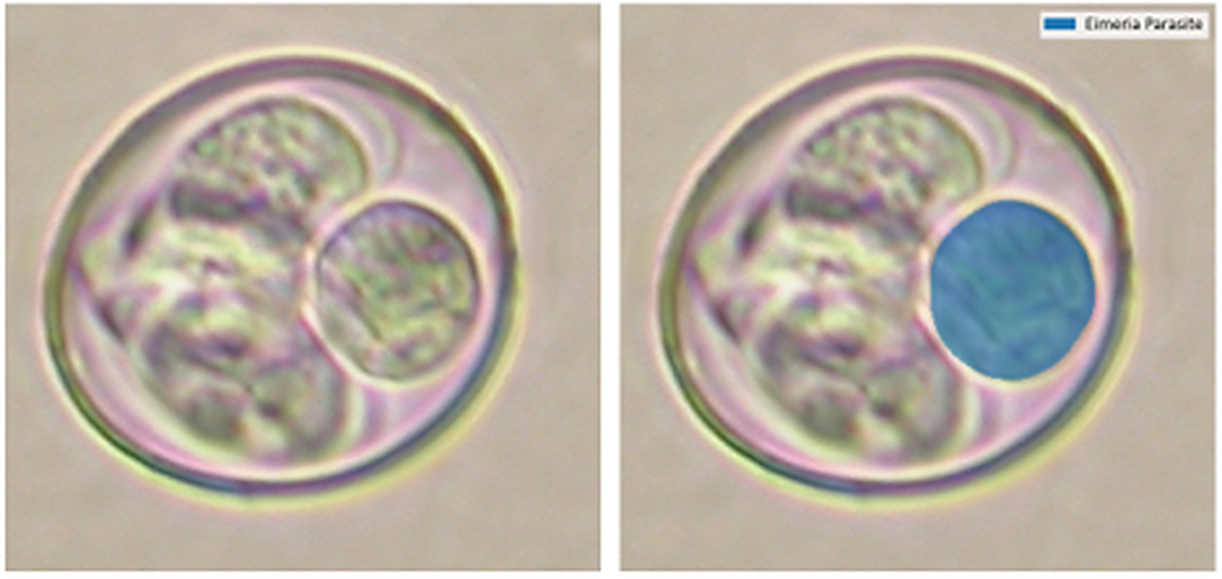}
    \caption{\textit{E. praecox}}
\end{subfigure}
\caption{Sample BiomedParse segmentation results. In each panel the left image is the input micrograph and the right image is the predicted segmentation (blue overlay) for the prompt ``Eimeria Parasite.''}
\label{fig:parasite_segmentation}
\end{figure}

The BiomedParse model was tested for segmenting images of the \textit{Eimeria} parasite, with the prompt ``Eimeria Parasite'' alongside images of different \textit{Eimeria} classes being fed into the model for segmentation. The results of the segmentation can be observed in Fig.~\ref{fig:parasite_segmentation}, which displays the input images alongside the model's output for different \textit{Eimeria} parasite classes.

As can be observed in the results, the model struggles to properly segment all of the parasite, but it still manages to segment some components of the parasites with varying degrees of success, as observed in the \textit{E. maxima} and \textit{E. praecox} samples for example. This suggests that with further fine-tuning on this dataset, which requires the manual creation of ground truth masks, the model can potentially segment \textit{Eimeria} parasite image samples with higher accuracy.

\subsection{Computational Cost and Inference Time}
The evaluated models differ markedly in computational profile. Gemini was accessed through Google's hosted API, requiring no local GPU, but each request incurs seconds of upload and remote-inference latency and is constrained by API rate limits, the exponential-backoff retries, and a per-call cost. Processing all 4,225 images was therefore time- and cost-bound. In contrast, specialized CNN classifiers run locally and are far more efficient. ResTFG, for example, classifies at 256 FPS with only 1.95M parameters on a single consumer GPU \cite{He2023}, while BiomedParse, a large foundation model, needs substantial GPU memory per segmentation. For high-throughput on-site diagnostics, lightweight local models are therefore far more efficient, while multimodal LLMs are more useful for flexible reasoning and report generation than for raw classification throughput. A hardware-controlled benchmark of latency, throughput, and cost remains as important future work.

\section{Conclusion}
This study evaluated Google Gemini for \textit{Eimeria} classification and pathology-report generation and explored BiomedParse for parasite segmentation. Classification was unreliable, with a strong default bias toward \textit{E. tenella}, showing that current prompt-based LLM classification is insufficient for species-level diagnosis when visual differences are subtle. The generated pathology reports were coherent and clinically plausible but require expert veterinary validation before any practical use, and BiomedParse segmented only parts of the oocysts, indicating a need for ground-truth masks and task-specific fine-tuning.

Overall, multimodal LLMs show promise as supportive tools for parasite image interpretation and reporting but are not yet suitable for standalone diagnostic use. Future work will focus on dataset expansion, model fine-tuning, improved segmentation annotation, and expert-based evaluation of the generated reports and treatment recommendations.

\bibliographystyle{IEEEtran}
\bibliography{bibfile}

\end{document}